\documentclass[conference]{IEEEtran}
\IEEEoverridecommandlockouts
\usepackage{amsmath,amssymb,amsfonts}
\usepackage{algorithmic}
\usepackage{graphicx}
\usepackage{textcomp}
\usepackage{xcolor}
\usepackage{multirow}
\usepackage{tabularx}
\usepackage{booktabs}
\usepackage{float}
\usepackage[table]{xcolor}
\usepackage{comment}

\definecolor{lightblue}{RGB}{200, 230, 255}

\def\BibTeX{{\rm B\kern-.05em{\sc i\kern-.025em b}\kern-.08em
    T\kern-.1667em\lower.7ex\hbox{E}\kern-.125emX}}

\begin{document}

\title{JitTrack: Onboard Multi-Object Tracking Against Viewpoint Jitter for Agile UAVs}

\author{
  \IEEEauthorblockN{Yachun Shan}
  \IEEEauthorblockA{%
    The Department of Advanced Manufacturing and Robotics,\\
    Peking University, Beijing, China\\
    2501213215@stu.pku.edu.cn
  }
  \and
  \IEEEauthorblockN{Feitian Zhang\IEEEauthorrefmark{1}}
  \IEEEauthorblockA{%
    The Department of Advanced Manufacturing and Robotics,\\
    Peking University, Beijing, China\\
    feitian@pku.edu.cn
  }
  \thanks{* Corresponding author}   % ✅ 放在这里（仍在 \author{} 内部）
}

\maketitle

\begin{abstract}
Multi-object tracking (MOT) onboard agile unmanned aerial vehicles (UAVs) remains challenging due to severe viewpoint jitter induced by camera ego-motion. Rapid attitude changes during flight often lead to significant target displacement across frames, causing inaccurate target association and degraded tracking performance. Existing UAV MOT methods are primarily evaluated on offline benchmarks and seldom address the practical requirements of real-world onboard deployment, including robustness to camera motion and active target following. To address these challenges, we propose JitTrack, an active onboard multi-object tracking framework that accommodates drone dynamics and camera ego-motion. Built upon a query-based transformer tracker, JitTrack introduces semantic refinement to improve the detection of emerging targets, motion-aware query rectification to compensate for target misalignment caused by viewpoint jitter, and a motion-inspired denoising training strategy that simulates camera motion patterns  for robust supervision. Furthermore, we develop a perception-planning-control closed-loop tracking pipeline  for real-world deployment, enabling collision-free and physically feasible target following on agile UAVs. Extensive experiments on public UAV MOT benchmarks demonstrate consistent improvements over the baseline method, while real-world flight experiments validate the effectiveness and practicality of JitTrack for robust onboard visual tracking under viewpoint jitter.
\end{abstract}


\section{Introduction}
Multi-object tracking (MOT) is a fundamental capability for embodied intelligence systems, with applications ranging from autonomous driving \cite{hu2023planning, 11018397, 11079959} and aerial surveillance \cite{oh2011large} to human-robot interaction \cite{bhatnagar2022behave}. Compared with conventional vision-based MOT with static cameras \cite{milan2016mot16, dendorfer2020mot20, sun2022dancetrack}, UAV-based MOT introduces additional challenges due to the strong coupling between camera motion and object motion. During agile flight, rapid changes in UAV attitude  induce significant viewpoint variations, causing abrupt target displacement and appearance changes across consecutive frames. Such disturbances severely affect target association and may lead to identity switches, especially for onboard systems with limited computation resources.

\begin{figure}[t]
\centering
\includegraphics[width=1\linewidth]{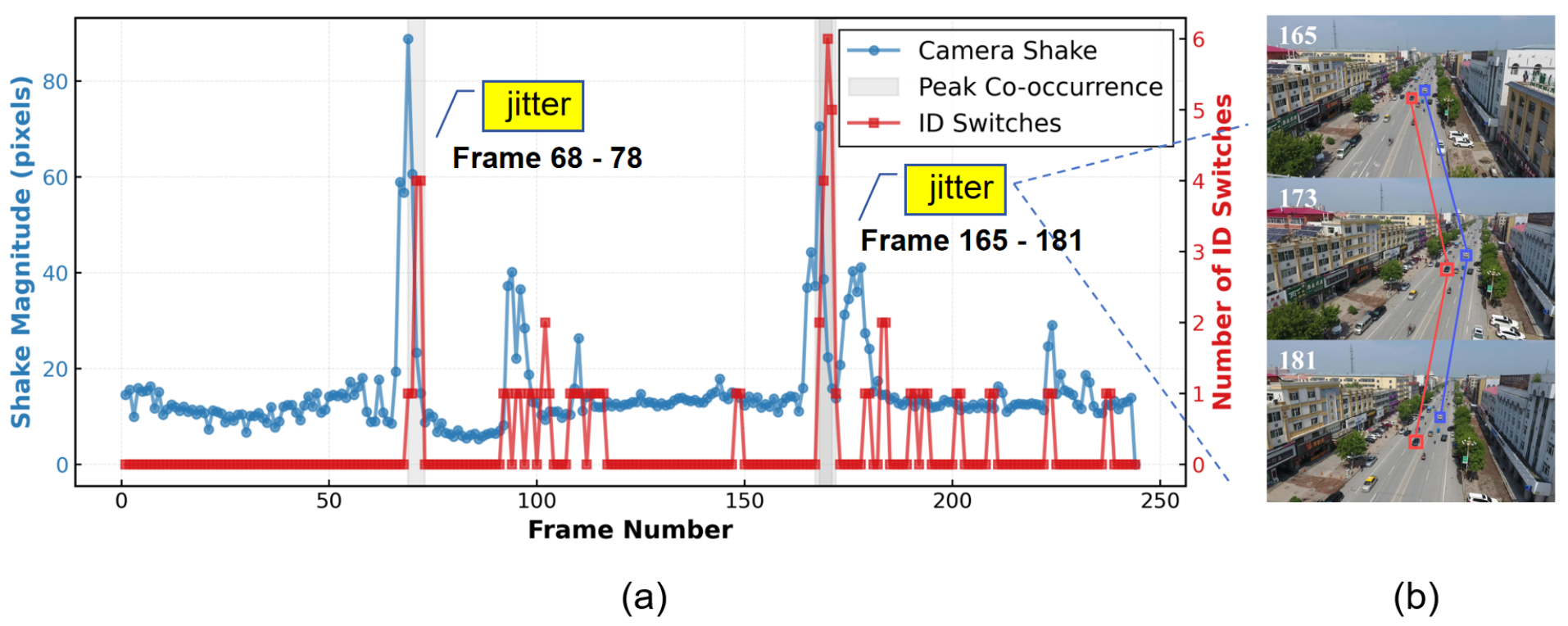}
\caption{\label{fig:jitter_ids}Influence of UAV camera jitter on multi-object tracking.
(a) Distribution of identity switches and camera shake intensity across frames from VisDrone2019 dataset. 
Identity switches frequently occur during periods of intensive camera motion, indicating that viewpoint variations degrade temporal association. (b) Visualization of UAV-induced viewpoint jitter. Rapid yaw motion introduces significant target displacement in the image plane, resulting in spatial inconsistency between consecutive frames and increased association ambiguity.
}
\end{figure}

Existing UAV MOT benchmarks and methods have mainly focused on challenges including small object scale, dense targets, and appearance variations \cite{zhu2021detection,du2018unmanned,zhang2019eye,liu2022multi}. However, most existing evaluations are conducted on offline video sequences, where camera trajectories are relatively smooth. The influence of dynamic UAV motion during real-world onboard deployment remains insufficiently explored. As illustrated in Fig.~\ref{fig:jitter_ids}, intensive camera jitter introduces large target displacement between adjacent frames, significantly increasing association ambiguity.

To improve robustness against camera motion, existing approaches typically rely on explicit motion compensation or additional motion modeling modules \cite{aharon2022bot,yi2024ucmctrack,zhang2019eye}. However, these methods often treat camera motion as an external factor and require separate estimation procedures, which increase system complexity and may become unreliable under rapid or irregular UAV maneuvers. Moreover, existing UAV MOT approaches mainly focus on  perception accuracy and rarely consider the closed-loop interaction among tracking, planning, and control required for persistent target following on physical platforms.

Recently, transformer-based MOT trackers have demonstrated strong temporal modeling capability through query-based object representation  \cite{meinhardt2022trackformer,zeng2022motr,gao2023memotr,roar2025}.  By propagating object queries across frames, these methods provide an effective framework for long-term tracking. However, their performance strongly depends on the temporal consistency of query representations. Under severe camera ego-motion, propagated queries may become spatially misaligned with current observations, resulting in degraded association performance. Therefore, improving the motion robustness of query-based trackers is essential for agile UAV deployment.

To address these challenges, we propose JitTrack, an active onboard multi-object tracking framework designed for robust tracking under viewpoint jitter. Instead of explicitly estimating and compensating camera motion, JitTrack enables the tracker to learn motion-aware representations. Specifically, we introduce three complementary strategies: (1) semantic query refinement, which improves the detection query representation by incorporating image-level information; (2) motion-aware query rectification, which adaptively adjust track queries according to global motion consistency; and (3) motion-inspired denoising training, which simulates UAV-induced motion patterns during training to improve robustness against target displacement. Furthermore, we establish a perception-planning-control closed-loop tracking pipeline and deploy JitTrack on a physical UAV platform for active target following.

The main contributions of our work are threefold:   
\begin{itemize}
\item We propose JitTrack, a motion-aware query-based MOT framework that improves robustness against viewpoint jitter caused by UAV camera ego-motion.
\item We introduce semantic query refinement, motion-aware query rectification, and motion-inspired denoising training strategies to enhance tracking robustness under dynamic flight conditions.
\item We develop a closed-loop onboard tracking system integrating perception, planning, and control, and validate the proposed method through extensive benchmark evaluations and real-world flight experiments.
\end{itemize}

\section{Related Work}

\subsection{UAV-Based Multi-Object Tracking}

UAV-based MOT has attracted increasing attention due to its importance in aerial perception applications. Existing UAV MOT benchmarks, such as UAVDT and VisDrone2019-MOT, provide large-scale datasets for evaluating tracking algorithms under aerial viewpoints \cite{du2018unmanned,zhu2021detection}. Recent approaches have investigated improved feature representation, temporal modeling, and association strategies to address challenges including scale variation, viewpoint changes, and complex backgrounds \cite{zhang2019eye,liu2022multi}. However, these methods mainly focus on improving tracking accuracy under predefined datasets and do not explicitly address severe camera ego-motion caused by agile UAV maneuvers. Although some onboard perception systems have investigated dynamic obstacle tracking on UAV platforms, their primary focus remains on navigation and collision avoidance rather than multi-object identity preservation under camera ego-motion \cite{10323166}.

\subsection{Transformer-Based Multi-Object Tracking}
Transformer-based MOT has recently emerged as a promising paradigm by formulating tracking as an end-to-end sequence prediction problem. DETR introduced object queries for transformer-based object detection, providing the foundation for subsequent query-driven tracking frameworks \cite{carion2020end}. TrackFormer extended this idea to MOT by propagating track queries across video frames to maintain object identities without explicit data association \cite{meinhardt2022trackformer}. MOTR further improved query propagation through track query updating and temporal modeling, enabling long-term tracking in complex video sequences \cite{zeng2022motr}. More recent approaches have explored memory mechanisms and enhanced temporal interaction to improve query-based tracking performance \cite{gao2023memotr}.

Despite their effectiveness, existing query-based trackers generally assume sufficient temporal consistency between adjacent frames. The propagated queries are highly dependent on previous observations and may become unreliable when objects undergo large displacement or abrupt appearance changes. Such limitations motivate the development of motion-aware query adaptation strategies for dynamic robotic platforms.

\subsection{Camera Motion Compensation for Multi-Object Tracking}

Tracking with moving cameras remains challenging due to the coupling between camera motion and object motion. Camera motion compensation (CMC) methods have been widely investigated to reduce the influence of global image motion by estimating transformations between consecutive frames. For example, BoT-SORT integrates camera motion compensation with appearance and motion cues to improve tracking robustness in dynamic scenes \cite{aharon2022bot}. UCMCTrack further investigates unified camera motion compensation strategies by explicitly modeling camera-induced motion \cite{yi2024ucmctrack}. Wang et al. introduces a proposal-based collaborative re-detection mechanism to address large displacement in low frame rate single-object tracking \cite{10697256}. Other approaches exploit geometric constraints and motion representations to improve tracking under camera movement \cite{zhang2019eye}.

Although effective, existing CMC approaches generally introduce independent motion estimation modules before tracking. Such explicit compensation pipelines increase computational complexity and may become unreliable under rapid or irregular camera motion. In contrast, learning motion-aware representations directly within the tracking framework provides a more compact solution for onboard robotic systems.

Meanwhile, query enhancement and denoising strategies have been explored to improve transformer optimization and detection robustness. DETR variants have introduced improved query initialization and denoising mechanisms to provide stronger supervision during training. For example, DN-DETR employs noisy label queries to accelerate convergence and improve transformer detection performance\cite{li2022dn}. However, existing query enhancement methods mainly focus on detection optimization and training stability, without explicitly considering motion disturbances caused by dynamic platforms.

\begin{figure*}[t!]
\centering
\includegraphics[width=\textwidth]{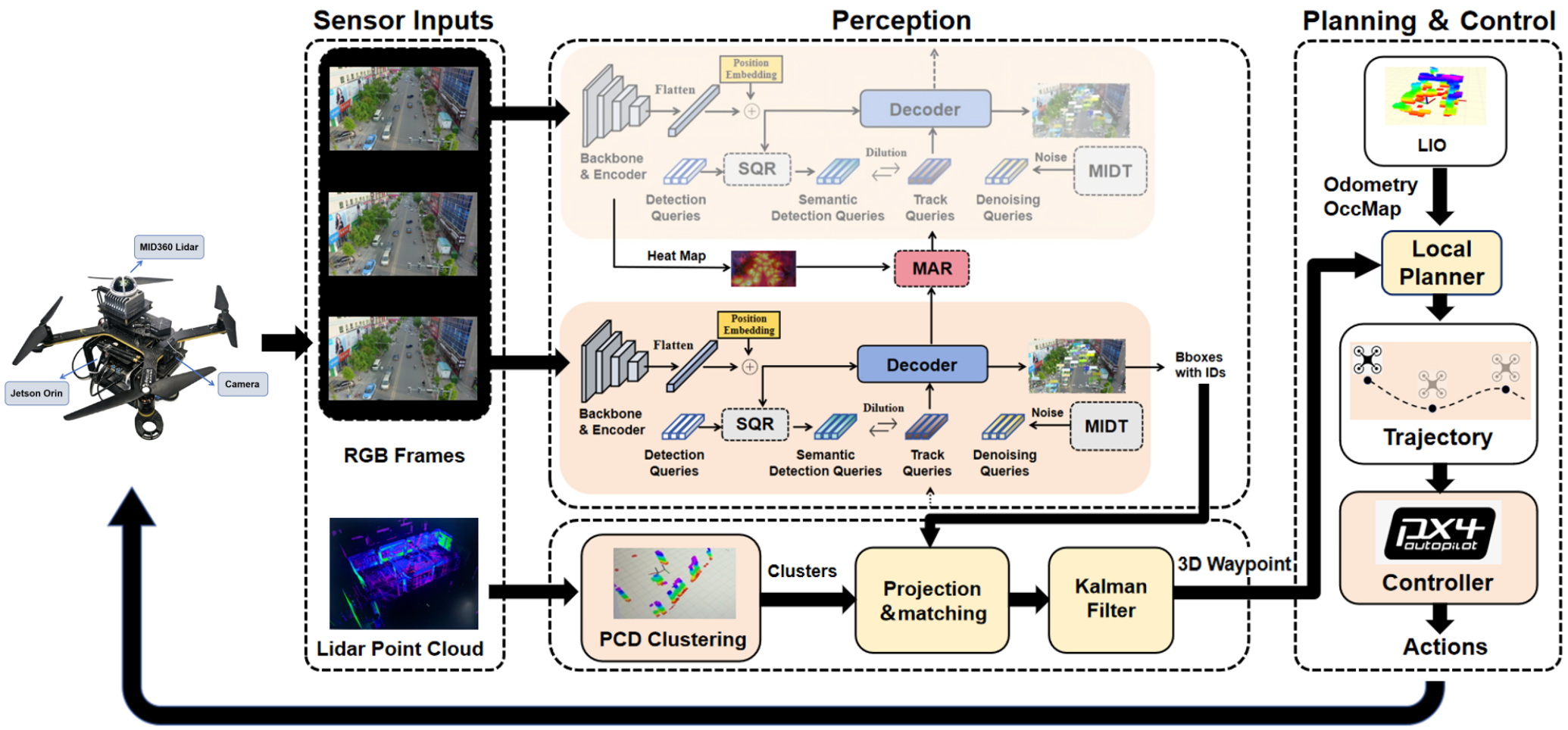}
\caption{\label{fig:main_graph}Overview of the proposed JitTrack framework. During training, JitTrack receives three types of queries: semantic detection queries refined by the semantic query refinement module, motion-rectified track queries compensated by the motion-aware query rectification module, and motion-inspired denoising queries generated from jitter-perturbed ground-truth targets. During inference, denoising queries are removed, and the refined tracking results are integrated with the onboard perception-planning-control pipeline for active UAV target following.}
\end{figure*}

Different from previous approaches, this work investigates motion-aware query learning for onboard UAV MOT under viewpoint jitter. JitTrack integrates semantic query refinement, motion-aware query rectification, and motion-inspired denoising training into a unified transformer tracking framework, enabling robust tracking without explicit camera motion estimation modules.

\section{Method}

As illustrated in Fig.~\ref{fig:main_graph}, JitTrack is a motion-aware query-based multi-object tracking framework designed for robust onboard UAV tracking under viewpoint jitter. 
Unlike conventional approaches that explicitly estimate camera motion before tracking, JitTrack improves robustness by enhancing the temporal consistency of query representations affected by camera ego-motion.

Specifically, the proposed framework consists of three complementary components. First, the semantic query refinement module enhances initialized detection queries by incorporating image-level semantic information, alleviating the imbalance between newly initialized queries and propagated track queries. 
Second, the motion-aware query rectification module exploits global motion consistency to compensate query position shifts caused by camera ego-motion. 
Third, the motion-inspired denoising training strategy introduces UAV-motion-related perturbations during training, enabling the tracker to learn robust localization under viewpoint jitter.
During inference, only detection and track queries are retained. The predicted tracking results are further fused with onboard sensing information to support downstream trajectory generation and active target following.

\begin{figure}[t!]
\centering
\includegraphics[width=1\linewidth]{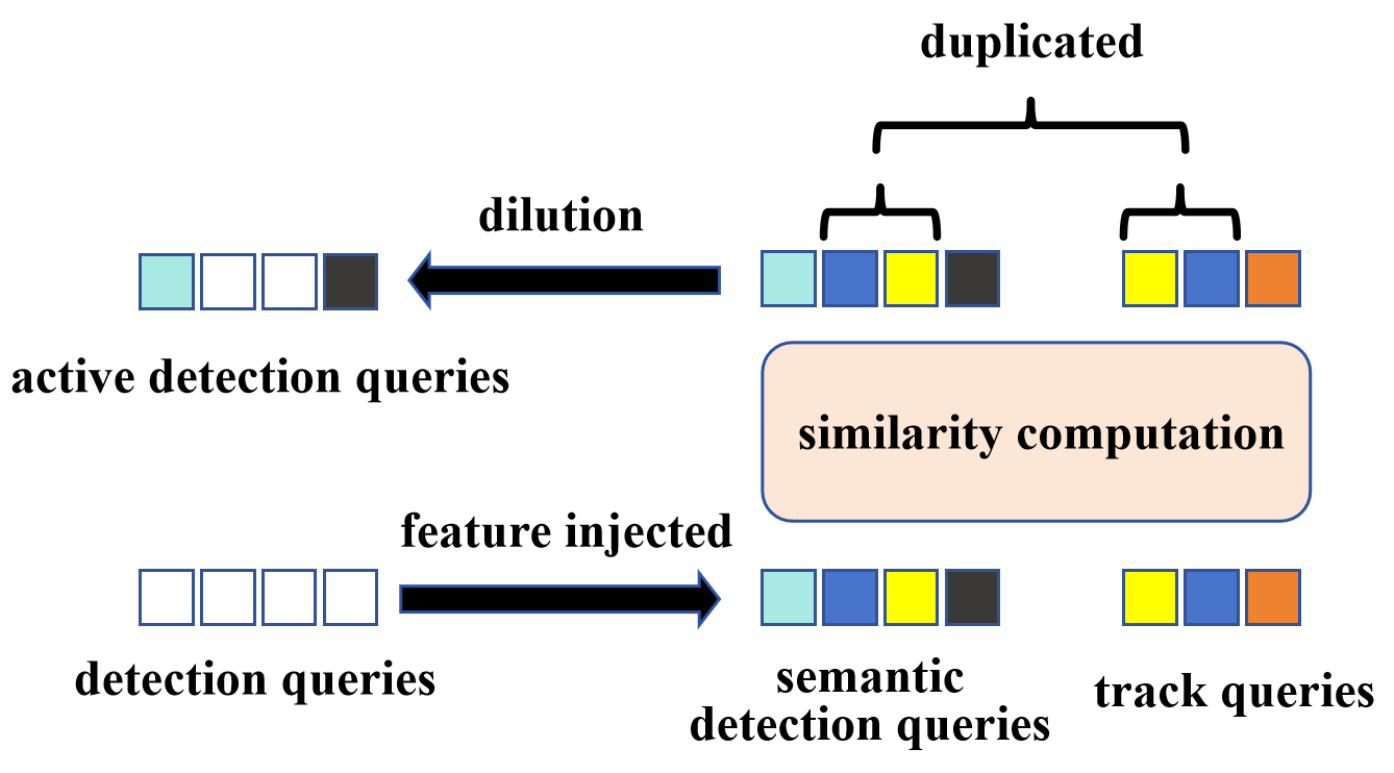}
\caption{\label{fig:SQR}Architecture of the semantic query refinement module. Image features are injected into initialized detection queries, while similarity-based dilution suppresses redundant responses caused by existing track queries.}
\end{figure}

\subsection{Semantic Query Refinement}

Query-based MOT trackers propagate track queries across frames to preserve target identities. However, newly initialized detection queries usually contain weaker semantic information than propagated track queries, causing the latter to dominate the self-attention interaction and suppress emerging targets  \cite{gao2023memotr, yan2023bridging, yu2023motrv3}. 

To address this imbalance, we introduce a semantic query refinement module that injects image features into the detection queries before decoder processing. Instead of treating detection queries as fixed learnable embeddings, the proposed module adaptively enriches them with current-frame visual information, improving their ability to discover newly appearing targets.

Given initialized detection queries $Q_\mathrm{d}$ and encoded image features $F$, semantic-enhanced detection queries are obtained through cross- and self-attention $Q_\mathrm{d}^\mathrm{sem}=\mathrm{SelfAttn}(\mathrm{CrossAttn}(Q_\mathrm{d},F))$. Although semantic enrichment improves target discovery, it may introduce duplicate responses for targets already represented by track queries. Therefore, we introduce a similarity-aware dilution mechanism, as illustrated in Fig. \ref{fig:SQR}. The similarity weight is calculated as $W_\mathrm{dil}=\mathrm{Sigmoid}(\mathrm{Mean}(Q_\mathrm{d}^\mathrm{sem}Q_\mathrm{t}^\mathrm{T}))$ where $Q_\mathrm{t}$ denotes propagated track queries. The refined detection queries are then obtained as
\begin{equation}
\tilde{Q}_\mathrm{d}^\mathrm{sem} = Q_\mathrm{d}^\mathrm{sem}(1-\lambda W_\mathrm{dil}),
\end{equation}
where $\lambda$ adjusts the dilution strength. This operation suppresses redundant responses while preserving sensitivity to newly appearing targets.

\begin{figure}[t]
\centering
\includegraphics[width=1\linewidth]{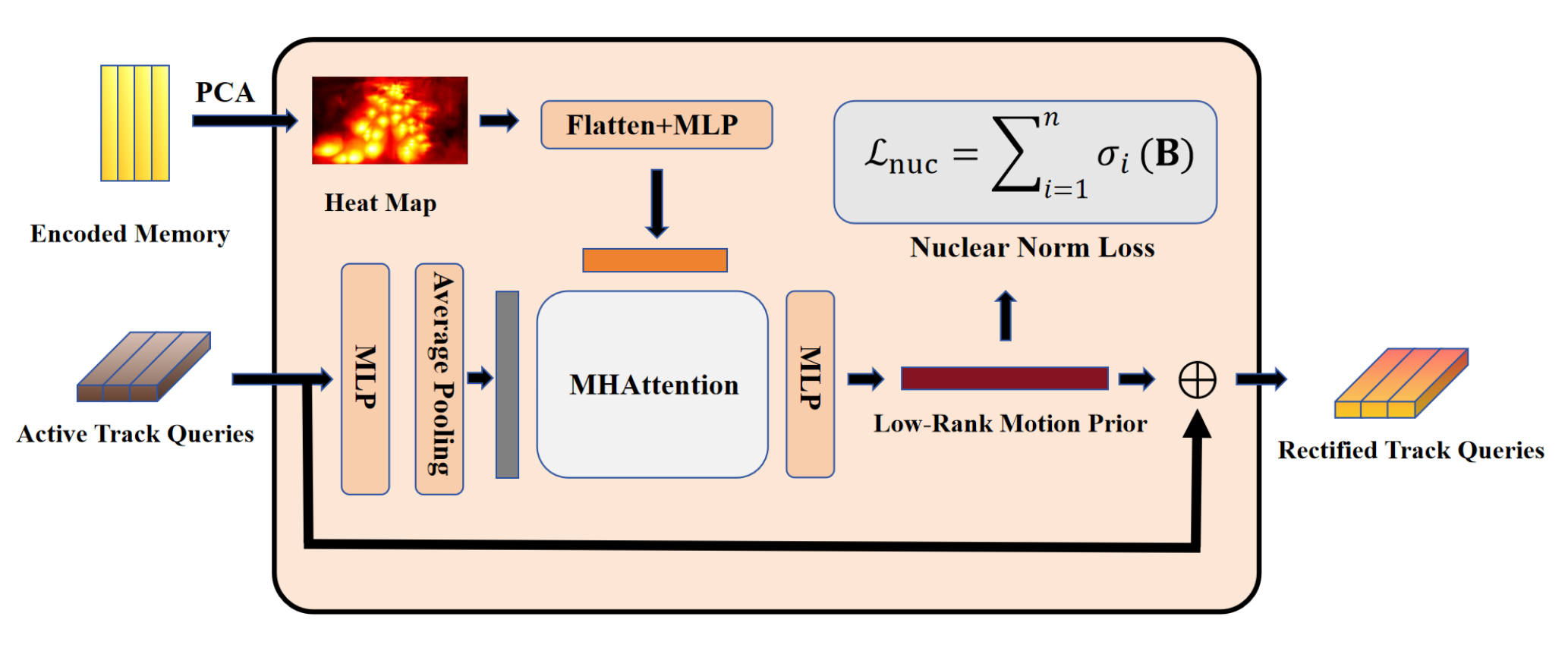}
\caption{\label{fig:MAR}Architecture of the motion-aware query rectification module. A heatmap-based spatial prior guides the estimation of a unified motion bias, which compensates query position offsets induced by camera ego-motion.}
\end{figure}

\subsection{Motion-Aware Query Rectification}

Camera ego-motion introduces global target displacement between adjacent frames, causing spatial inconsistency between propagated track queries and current observations. Existing query-based trackers mainly rely on attention mechanisms for implicit matching, which may become unreliable when severe viewpoint jitter occurs.

To address this issue, we propose a motion-aware query rectification module that explicitly models global motion consistency. 
The key observation is that camera-induced displacement exhibits a low-dimensional structure because background motion generated by camera rotation is approximated by a low-degree-of-freedom transformation.

As illustrated in Fig. \ref{fig:MAR}, the module first extracts a spatial motion prior from current-frame encoder features. The previous-frame track queries are projected into a shared latent space, and a unified motion bias is estimated through motion-conditioned multi-head attention $B_\mathrm{m} = \mathrm{MHA}(Q_\mathrm{t},H_\mathrm{t})$, where $H_\mathrm{t}$ represents the heatmap-based spatial prior and $B_\mathrm{m}$ denotes the estimated motion bias.
The position embeddings of track queries are then rectified as
\begin{equation}
P_\mathrm{t}^{'}=P_\mathrm{t}+B_\mathrm{m} ,
\end{equation}
where $P_\mathrm{t}$ and $P_\mathrm{t}^{'}$ represent the original and rectified query positions, respectively.

Different from independent camera motion compensation methods, the proposed rectification module directly updates query representations inside the tracker. Moreover, to enforce the global consistency of camera motion, a low-rank constraint is imposed on the motion bias using the nuclear norm loss $\mathcal{L}_\mathrm{nuc}=||B_\mathrm{m}||_*$.

\subsection{Motion-Inspired Denoising Training}

Denoising training has been shown effective for improving transformer optimization by reconstructing noisy targets \cite{li2022dn,fu2023denoising}. However, conventional denoising strategies mainly consider random localization perturbations and do not explicitly model camera-induced motion.

\begin{figure}[t!]
\centering
\includegraphics[width=0.9\linewidth]{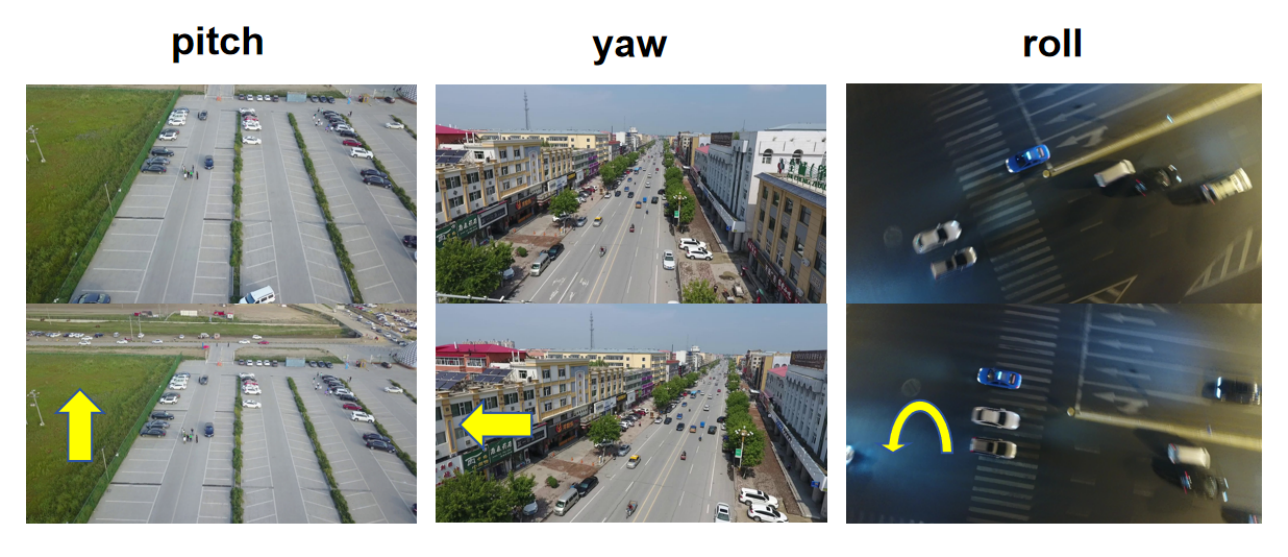}
\caption{\label{fig:three_modes}Three motion modes of a drone camera. For distant targets and small camera viewpoint changes, the induced target displacement is modeled as a linear combination of translation- and rotation-induced motion components.}
\end{figure}

To improve robustness against UAV viewpoint jitter, we introduce motion-inspired denoising training. The camera motion of agile UAVs is approximated by three dominant image-plane transformations: horizontal translation, vertical translation, and rotation around the optical axis, as illustrated in Fig.~\ref{fig:three_modes}. Therefore, we generate denoising queries by perturbing ground-truth bounding boxes according to yaw, pitch, and roll motions.

For yaw and pitch disturbances, the bounding box center is shifted as
\begin{equation}
B_\mathrm{yaw} = (x+(1+\epsilon)\lambda_1 w,y,w,h),
\end{equation}
\begin{equation}
B_\mathrm{pitch} = (x,y+(1+\epsilon)\lambda_2 h,w,h),
\end{equation}
where $\lambda_1$ and $\lambda_2$ represent horizontal and vertical motion offsets.

For roll motion, the bounding box center is rotated around the image center, where the rotation angle $\theta$ is randomly sampled to construct the rotation matrix $R(\theta)$, i.e.,
\begin{equation}
\begin{bmatrix}
x'\\y'
\end{bmatrix}
=
R(\theta)
\begin{bmatrix}
x-W/2\\y-H/2
\end{bmatrix}
+
\begin{bmatrix}
W/2\\H/2
\end{bmatrix}
+\epsilon .
\end{equation}
The generated denoising queries are combined with detection and track queries $Q=\{Q_\mathrm{dn},Q_\mathrm{d},Q_\mathrm{t}\}$.

\begin{figure}[htbp]
\centering
\includegraphics[width=0.9\linewidth]{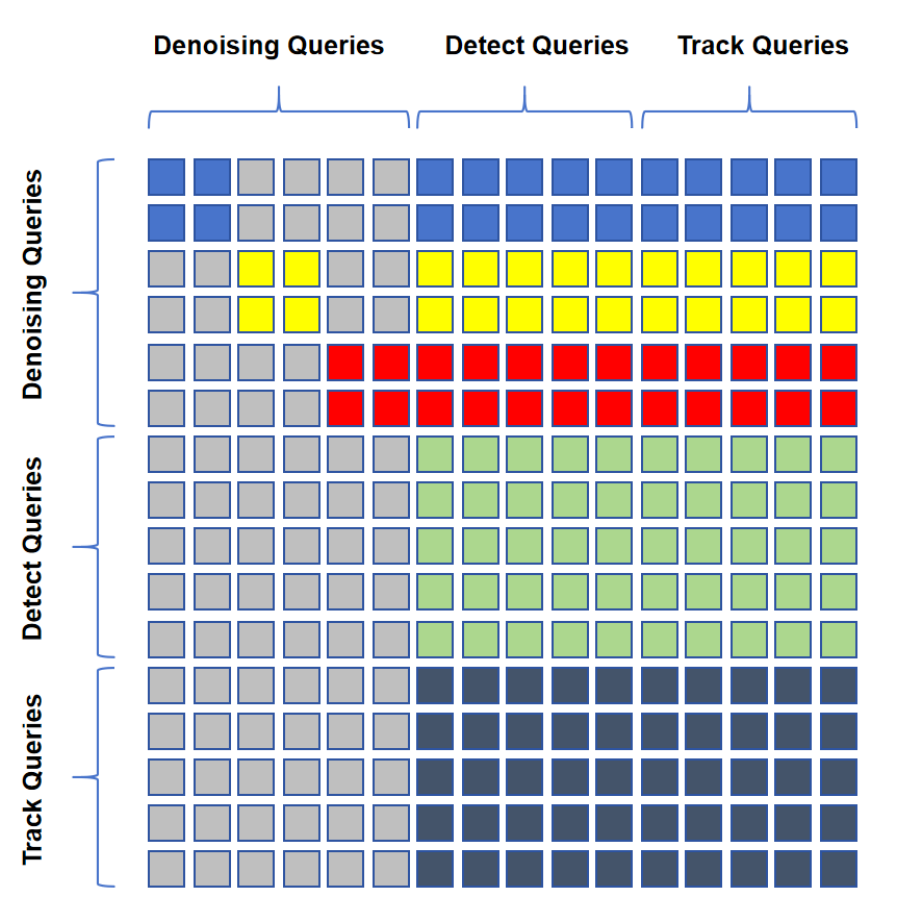}
\caption{\label{fig:query_set_mask} Attention mask for the joint query set. The masked regions prevent detection and track queries from accessing ground-truth information contained in denoising queries during training.
}
\end{figure}

To avoid information leakage from ground-truth denoising queries, an attention mask is applied during self-attention. As illustrated in Fig.~\ref{fig:query_set_mask}, detection and track queries are prevented from accessing denoising queries.

\subsection{Optimization Objective}

The overall objective consists of three components: the jitter-aware drift loss, the motion regularization loss, and the denoising reconstruction loss.

The proposed drift loss increases the contribution of samples with large temporal displacement, defined as
\begin{equation}
\mathcal{L}_\mathrm{drift}
=
\lambda_\mathrm{cls}\mathcal{L}_\mathrm{cls}
+
W_\mathrm{drift}
(
\lambda_\mathrm{bbox}\mathcal{L}_\mathrm{bbox}
+
\lambda_\mathrm{giou}\mathcal{L}_\mathrm{giou}
).
\end{equation}
Here, $\mathcal{L}_\mathrm{bbox}$ denotes the bounding box regression loss,  $\mathcal{L}_\mathrm{giou}$ represents the generalized intersection-over-union loss, and $W_\mathrm{drift}$ is the drift weight.

The nuclear norm regularization constrains the estimated motion bias to maintain a low-rank structure, i.e.,
\begin{equation}
\mathcal{L}_\mathrm{nuc}=||B_\mathrm{m}||_* .
\end{equation}

The denoising reconstruction loss supervises the recovery of jitter-corrupted queries, defined as L2 loss between the predicted
reconstructed boxes $b_i$ and the corresponding ground-truth boxes $b_i^\mathrm{gt}$, i.e.,
\begin{equation}
\mathcal{L}_\mathrm{dn}
=
\frac{1}{N}
\sum_i
||b_i-b_i^\mathrm{gt}||_2^2 .
\end{equation}

The final training objective is:
\begin{equation}
\mathcal{L} = \mathcal{L}_{\mathrm{drift}} + \mathcal{L}_{\mathrm{nuc}} + \mathcal{L}_{\mathrm{dn}}
\end{equation}

\begin{comment}
\subsection{Real-World Deployment}

To bridge the gap between MOT and real-world visual tracking tasks, we propose a perception-planning-control closed-loop tracking pipeline for onboard platform deployment, as illustrated in Fig.~\ref{fig:main_graph}. The perception module processes RGB frames and LiDAR point clouds asynchronously. Visual tracking results are fused with point-cloud clusters via a Kalman filter to locate targets positions as stable 3D waypoints, which are then fed into the Fast-Planner \cite{zhou2019robust}. The planner computes collision-free trajectories using the odometry from FAST-LIO \cite{xu2021fastliofastrobustlidarinertial}. Finally, the planned trajectories are executed by a PX4 controller, ensuring a safe and decoupled control loop.
\end{comment}

\begin{table*}[htbp]
\caption{Comparison with state-of-the-art MOT methods on UAV benchmarks. 
JitTrack achieves improved identity preservation and robustness under dynamic aerial viewpoints.
}
\centering
\small
\setlength{\tabcolsep}{5pt}
\begin{tabular}{|l|c|c|c|c|c|c|c|c|c|}
\hline
\textbf{Dataset} & \textbf{Method} & \textbf{MOTA↑(\%)} & \textbf{MOTP↑(\%)} & \textbf{IDF1↑(\%)} & \textbf{MT↑} & \textbf{ML↓} & \textbf{FP↓} & \textbf{FN↓} & \textbf{IDs↓} \\ \hline
\multirow{8}{*}{\textbf{VisDrone2019}} 
& SiamMOT & 31.9 & 73.5 & 48.3 & - & - & 24123 & 142303 & 862 \\ 
& ByteTrack & 25.1 & 72.4 & 40.8 & 446 & 1099 & 34044 & 194984 & 1590 \\ 
& UAVMOT & 36.1 & 74.2 & 51.0 & 520 & 574 & 27983 & 115925 & 2775 \\ 
& FOLT & 42.1 & \textbf{77.6} & 56.9 & - & - & \textbf{24105} & 107630 & \textbf{800} \\ 
& DroneMOT & \textbf{43.7} & 71.4 & \textbf{58.6} & \textbf{689} & \textbf{397} & 41998 & \textbf{86177} & 1112 \\ 
& \cellcolor{lightblue}MOTR & \cellcolor{lightblue}22.8 & \cellcolor{lightblue}72.8 & \cellcolor{lightblue}41.4 & \cellcolor{lightblue}272 & \cellcolor{lightblue}825 & \cellcolor{lightblue}28407 & \cellcolor{lightblue}147937 & \cellcolor{lightblue}959 \\ 
& \cellcolor{lightblue}TrackFormer & \cellcolor{lightblue}25.0 & \cellcolor{lightblue}73.9 & \cellcolor{lightblue}30.5 & \cellcolor{lightblue}385 & \cellcolor{lightblue}770 & \cellcolor{lightblue}25856 & \cellcolor{lightblue}141526 & \cellcolor{lightblue}4840 \\ 
& \cellcolor{lightblue}Ours & \cellcolor{lightblue}\textbf{38.4} & \cellcolor{lightblue}\textbf{75.9} & \cellcolor{lightblue}\textbf{53.1} & \cellcolor{lightblue}\textbf{641} & \cellcolor{lightblue}\textbf{488} & \cellcolor{lightblue}\textbf{21222} & \cellcolor{lightblue}\textbf{117422} & \cellcolor{lightblue}\textbf{907} \\ \hline
\multirow{6}{*}{\textbf{UAVDT}} 
& SiamMOT & 39.4 & 76.2 & 61.4 & - & - & 46903 & 176164 & 190 \\ 
& ByteTrack & 41.6 & 79.2 & 59.1 & - & - & \textbf{28819} & 189197 & 296 \\ 
& UAVMOT & 46.4 & 72.7 & 67.3 & 624 & 221 & 66352 & 115940 & 456 \\ 
& FOLT & 48.5 & \textbf{80.1} & 68.3 & - & - & 36429 & 155696 & 338 \\ 
& DroneMOT & \textbf{50.1} & 74.5 & \textbf{69.6} & 638 & \textbf{178} & 57411 & \textbf{112548} & \textbf{129} \\ 
& Ours & 46.7 & 74.6 & 68.1 & \textbf{678} & 325 & 45180 & 136285 & 212 \\ \hline
\end{tabular}
\label{tab:sota}
\end{table*}

\section{Experiments}
This section evaluates JitTrack through comprehensive experiments on public UAV MOT benchmarks and a real-world UAV active tracking system. We first compare JitTrack with representative state-of-the-art MOT approaches under challenging aerial viewpoints. Then, ablation studies are conducted to analyze the contribution of each proposed component. Finally, the proposed framework is validated through closed-loop active target tracking experiments on a physical UAV platform.

\subsection{Experimental Setup}
\subsubsection{Datasets and Evaluation Metrics}
We evaluate JitTrack on two widely used UAV MOT benchmarks, including VisDrone2019-MOT and UAVDT. 

The VisDrone2019 dataset \cite{zhu2021detection} contains 288 video clips consisting of 261,908 frames and 10,209 static images captured by various drone-mounted cameras. The dataset covers diverse locations, environments, object categories, and flight conditions, with more than 2.6 million manually annotated bounding boxes. In this work, we use the VisDrone2019-MOT subset for multi-object tracking, where the model is trained on VisDrone2019-MOT-train and evaluated on VisDrone2019-test-dev.

The UAVDT benchmark \cite{du2018unmanned} consists of 100 UAV-captured video sequences collected from urban environments, including squares, arterial streets, toll stations, highways, crossings, and T-junctions. The videos are recorded at 30~fps with a resolution of $1080 \times 540$. Following the standard protocol, we use UAV-benchmark-M, which contains 30 sequences for training and 20 sequences for evaluation.

Following previous MOT studies, we report standard tracking metrics, including Multiple Object Tracking Accuracy (MOTA), Higher Order Tracking Accuracy (HOTA), Identification F1-score (IDF1), and the number of identity switches (IDs). In addition, we analyze tracking robustness under different levels of camera jitter to evaluate the influence of viewpoint variation on identity preservation.

\subsubsection{Implementation Details}

JitTrack is implemented based on a transformer-based MOT framework and trained on four NVIDIA GeForce RTX 4090 GPUs for 100 epochs. The video clip sampler length is initially set to 2 and progressively increased to 3 and 4 after the 50th and 75th epochs, respectively, to enhance temporal modeling capability. During training, MOT-specific augmentations, including random resizing and random cropping, are adopted. 

The Adam optimizer is used with an initial learning rate of $2\times10^{-4}$, which is reduced by a factor of 10 after the 60th epoch. The batch size is set to 1 due to the high memory consumption of video-based MOT training. The backbone network is ResNet-50, and both the transformer encoder and decoder contain 6 layers with 8 attention heads. 

The number of initialized detection queries is set to 300. For the motion-inspired denoising training strategy, the random perturbation ratio $\varepsilon$ is set to 0.1, and the rotation disturbance $\theta$ is uniformly sampled from the range $[-\Theta, \Theta]$, where the maximum rotation angle $\Theta$ is set to $\pi/30$.

\subsection{Comparison With State-of-the-Art Methods}

Table~\ref{tab:sota} compares JitTrack with representative state-of-the-art MOT approaches on the VisDrone2019-MOT and UAVDT benchmarks. The compared methods include both tracking-by-detection approaches with explicit association modules and end-to-end query-based trackers.

On the VisDrone2019-MOT benchmark, JitTrack achieves 38.4\% MOTA, 75.9\% MOTP, and 53.1\% IDF1, outperforming the baseline by 15.6\%, 3.1\%, and 11.7\%, respectively.
Among end-to-end trackers without additional matching modules, JitTrack achieves the best overall performance, demonstrating the effectiveness of the proposed motion-aware query adaptation strategy for UAV-based MOT.
In particular, the improved IDF1 score indicates that JitTrack effectively preserves target identities under severe viewpoint variations, where propagated queries may suffer from spatial inconsistency caused by camera ego-motion.

On the UAVDT benchmark, JitTrack achieves 46.7\% MOTA, 74.6\% MOTP, and 68.1\% IDF1, outperforming several existing approaches and achieving competitive performance with tracking-by-detection methods. 
The consistent improvements on both benchmarks demonstrate that the proposed semantic query refinement, motion-aware query rectification, and motion-inspired denoising training strategies effectively enhance the robustness of end-to-end query-based trackers under dynamic UAV viewpoints.

Compared with conventional tracking-by-detection pipelines, JitTrack does not require additional detection or association modules. Instead, it improves temporal association through motion-aware query learning, providing a more compact solution for onboard UAV tracking systems.

\begin{table}[t!]
\caption{Ablation study of the proposed components.
MIDT denotes motion-inspired denoising training, SQR denotes semantic query refinement, and MAR denotes motion-aware query rectification.
}
\centering
\footnotesize
\setlength{\tabcolsep}{3pt}
\begin{tabular}{|c|c|c|c|c|c|c|}
\hline
\textbf{Baseline} & \textbf{MIDT} & \textbf{SQR} & \textbf{MAR} & \textbf{IDF1↑(\%)} & \textbf{MOTA↑(\%)} & \textbf{IDs↓} \\ \hline
$\checkmark$ &  &  &  & 41.4 & 22.8 & 959 \\
$\checkmark$ & $\checkmark$ & & & 45.7 & 28.4 & 1087 \\
$\checkmark$ & $\checkmark$ & $\checkmark$ & & 48.3 & 36.0 & 920 \\
$\checkmark$ & $\checkmark$ & $\checkmark$ & $\checkmark$ & 53.1 & 38.4 & 907 \\ \hline
\end{tabular}
\label{tab:ablation_study}
\end{table}

\begin{table}[t!]
\caption{Comparison of different denoising strategies. DTS introduces random Gaussian perturbations, while MIDT generates motion-inspired perturbations according to UAV camera motion patterns.
}
\centering
\small
\setlength{\tabcolsep}{4pt}
\begin{tabular}{|c|c|c|c|c|c|}
\hline
\textbf{Baseline} & \textbf{DTS} & \textbf{MIDT} & \textbf{MOTA↑(\%)} & \textbf{Rcll↑(\%)} & \textbf{IDs↓} \\ \hline
$\checkmark$ &  &  & 22.8 & 28.5 & 959 \\
$\checkmark$ & $\checkmark$ & & 25.2 & 36.0 & 1195 \\
$\checkmark$ & $\checkmark$ & $\checkmark$ & 28.4 & 41.4 & 1087 \\ \hline
\end{tabular}
\label{tab:noise_addition_methods}
\end{table}

\subsection{Ablation Studies}

To analyze the contribution of each proposed component, we conduct ablation studies by progressively introducing motion-inspired denoising training (MIDT), semantic query refinement (SQR), and motion-aware query rectification (MAR).

Table~\ref{tab:ablation_study} summarizes the results. The baseline represents the original query-based tracker without any proposed modifications.

By introducing MIDT, the tracker achieves 45.7\% IDF1 and 28.4\% MOTA, 4.3\% and 5.6\% higher than the baseline, demonstrating the effectiveness of incorporating UAV motion patterns into the denoising training process. 
Unlike conventional denoising strategies that introduce random perturbations, MIDT generates motion-aware disturbances that better simulate camera-induced target displacement, enabling the tracker to learn more robust query representations under viewpoint variations.

After adding SQR, the tracking performance is further improved by 2.6\% IDF1 and 7.6\% MOTA. 
This improvement indicates that incorporating image-level semantic information into detection queries enhances the representation of newly initialized targets and reduces the discrepancy between detection queries and propagated track queries.

\begin{figure*}[t!]
\centering
\includegraphics[width=\textwidth]{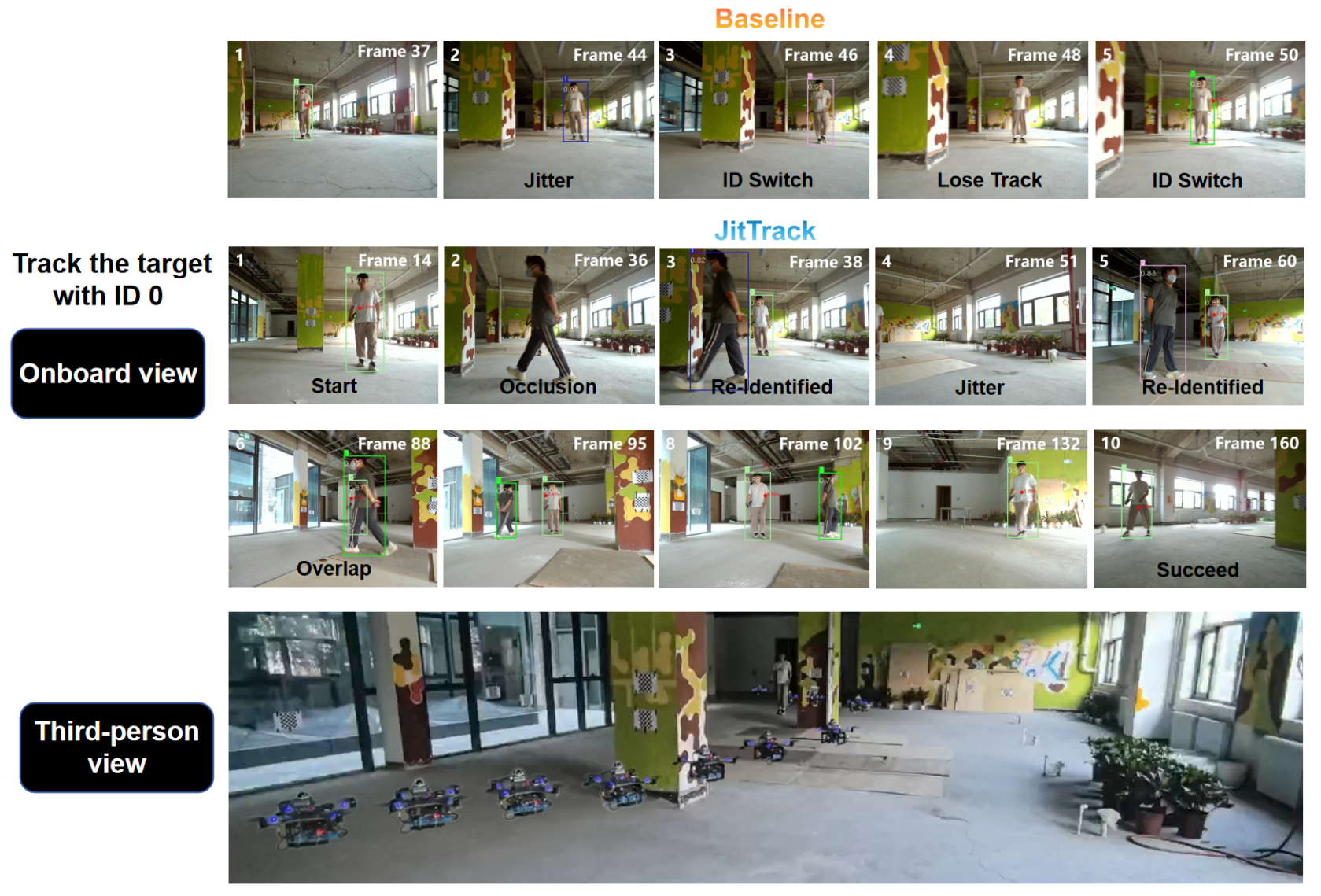}
\caption{\label{fig:real-world-test}Real-world UAV active tracking results. In this experiment, the UAV is tasked with tracking Target ID 0 while a distractor continuously walks back and forth across it to create repeated occlusions. Results demonstrate the robustness of our method compared with the baseline under severe viewpoint changes.
}
\end{figure*}

Finally, incorporating MAR achieves 53.1\% IDF1 and 38.4\% MOTA, particularly in identity-related metrics. 
Since camera ego-motion introduces global displacement between consecutive frames, propagated queries may become spatially inconsistent with current observations. 
MAR improves temporal association by adaptively rectifying query locations according to motion consistency, thereby reducing identity switches caused by viewpoint jitter.

\subsection{Analysis of Motion-Inspired Denoising Training}
To further investigate the effectiveness of MIDT, we compare different denoising strategies, including the conventional Denoising Training Strategy (DTS) and the proposed Motion-Inspired Denoising Training Strategy (MIDT), as shown in Table~\ref{tab:noise_addition_methods}.

The conventional DTS introduces random Gaussian noise to the position and size of bounding boxes, providing additional noisy supervision to improve transformer optimization. 
In contrast, MIDT generates perturbations according to UAV camera motion patterns, including viewpoint-induced displacement, to better reflect the target shifts observed during dynamic flight.

The results demonstrate that MIDT achieves better tracking performance than DTS. 
In particular, MIDT significantly improves the recall rate, indicating that motion-aware denoising enhances the model's capability to recover targets under large viewpoint variations. 
The improved tracking performance further verifies that simulating realistic UAV motion disturbances during training is beneficial for robust query learning.

It is worth noting that the baseline model may obtain fewer identity switches in some cases due to its inferior detection capability. 
Since missed detections reduce the possibility of generating identity changes, a lower ID switch value alone does not necessarily indicate better tracking performance. 
Therefore, identity preservation should be evaluated jointly with detection and association metrics, such as recall and IDF1.

\subsection{Real-World UAV Active Tracking}

\subsubsection{System Setup}

To validate the practical applicability of JitTrack beyond offline MOT benchmarks, we deploy the proposed tracker on a quadrotor UAV platform with a perception-planning-control closed-loop architecture. 

The perception module, running on an NVIDIA Jetson Orin NX 16GB onboard computer, asynchronously processes RGB images and MID360 LiDAR point clouds. The visual tracking results generated by JitTrack are associated with LiDAR-based target clusters and fused through a Kalman filter to estimate stable three-dimensional target positions. These target states are converted into navigation waypoints and provided to the Fast-Planner~\cite{zhou2019robust}, which generates collision-free trajectories according to the estimated target motion and environmental constraints. 

The planned trajectories are executed by a PX4-based flight controller, while the UAV state estimation is provided by FAST-LIO~\cite{xu2021fast}. This decoupled perception-planning-control architecture enables real-time target following under dynamic viewpoints and allows the proposed MOT method to be evaluated in a complete robotic deployment scenario.

\subsubsection{Active Tracking Results}

The UAV performs target following experiments under different flight trajectories with occlusion, including straight-line motion, turning maneuvers, and aggressive viewpoint changes.

Fig.~\ref{fig:real-world-test} presents representative tracking results. During rapid UAV rotations, conventional trackers frequently suffer from identity switches due to abrupt target displacement. In contrast, despite repeated target occlusions and camera jitter, JitTrack consistently maintains tracking of target ID 0 throughout the sequence, achieving zero ID switches for this target. 

Specifically, we evaluate the proposed method through 10 repeated experiments under identical challenging conditions. The baseline tracker fails to complete the full trajectory in all trials due to frequent identity switches, yielding a success rate of 0\%. In comparison, JitTrack completes the tracking task with a success rate of 90\%, and maintains zero ID switches throughout the successful trials. These results demonstrate that JitTrack not only excels on offline benchmarks but also provides reliable and robust perception for real-time robotic applications under dynamic viewpoints.

\section{Conclusion}

In this work, we proposed JitTrack, an end-to-end motion-aware multi-object tracking framework for UAV platforms under severe camera ego-motion. By improving query representation and motion adaptability, JitTrack enhances the robustness of query-based MOT against viewpoint-induced target displacement. Specifically, the proposed semantic query refinement improves the representation of detection and track queries, while the motion-aware query rectification adaptively compensates for motion-induced query misalignment. Furthermore, the motion-inspired denoising training strategy introduces realistic UAV motion disturbances during training, enabling the tracker to better handle camera jitter. Extensive experiments on UAV MOT benchmarks demonstrate that JitTrack achieves superior tracking performance compared with existing approaches, particularly in identity preservation under dynamic viewpoints. In addition, real-world UAV experiments validate the effectiveness of the proposed framework in a complete perception-planning-control closed-loop system. Future work will investigate more efficient motion-aware tracking strategies for resource-constrained autonomous platforms and more complex multi-agent aerial scenarios.

\bibliography{references}

\end{document}